%% file: 3590.tex
\documentclass[runningheads]{llncs}
\usepackage{graphicx}
\usepackage{comment}
\usepackage{amsmath,amssymb} 
\usepackage{color}

\usepackage{xcolor}
\usepackage{amssymb}
\usepackage{multicol}
\usepackage{multirow}
\usepackage{wrapfig}
\usepackage{soul}
\usepackage[caption=false]{subfig}
\usepackage{layouts}

\includecomment{COMMENT}    

\begin{document}
\pagestyle{headings}
\mainmatter
\def\ECCVSubNumber{3590}  

\title{Reparameterizing Convolutions for Incremental Multi-Task Learning without Task Interference} 

\titlerunning{Reparameterizing Convolutions for Multi-Task Learning}
%
\author{Menelaos Kanakis\inst{1} \and
David Bruggemann\inst{1} \and
Suman Saha\inst{1} \and \\
Stamatios Georgoulis\inst{1} \and
Anton Obukhov\inst{1} \and
Luc Van Gool\inst{1,2}}

%
\authorrunning{M. Kanakis et al.}
%
\institute{$^1$ ETH Zurich \hspace{1cm} $^2$ KU Leuven}


\maketitle

\input{text/abstract}

\input{text/intro}
\input{text/related_work}
\input{text/methodology}
\input{text/exp}
\input{text/conclusion}


\clearpage
%
%

\pagestyle{headings}
\def\ECCVSubNumber{3590}  

\title{Reparameterizing Convolutions for Incremental Multi-Task Learning without Task Interference\\(Supplementary Material)} 

\titlerunning{Reparameterizing Convolutions for Multi-Task Learning}
%
\author{Menelaos Kanakis\inst{1} \and
David Bruggemann\inst{1} \and
Suman Saha\inst{1} \and \\
Stamatios Georgoulis\inst{1} \and
Anton Obukhov\inst{1} \and
Luc Van Gool\inst{1,2}}
%

\authorrunning{M. Kanakis et al.}
%
\institute{$^1$ ETH Zurich \hspace{1cm} $^2$ KU Leuven}

\maketitle

\appendix
\input{text/sup_impl_details}
\input{text/sup_reparam}
\input{text/sup_baseline}
\input{text/sup_Exp_VGG}


\clearpage
%
%
\bibliographystyle{splncs04}
\bibliography{3590}

\end{document}

%% file: text/abstract.tex
\begin{abstract}

Multi-task networks are commonly utilized to alleviate the need for a large number of highly specialized single-task networks. However, two common challenges in developing multi-task models are often overlooked in literature. First, enabling the model to be inherently incremental, continuously incorporating information from new tasks without forgetting the previously learned ones (incremental learning). Second, eliminating adverse interactions amongst tasks, which has been shown to significantly degrade the single-task performance in a multi-task setup (task interference). In this paper, we show that both can be achieved simply by reparameterizing the convolutions of standard neural network architectures into a non-trainable shared part (filter bank) and task-specific parts (modulators), where each modulator has a fraction of the filter bank parameters. Thus, our reparameterization enables the model to learn new tasks without adversely affecting the performance of existing ones. The results of our ablation study attest the efficacy of the proposed reparameterization. Moreover, our method achieves state-of-the-art on two challenging multi-task learning benchmarks, PASCAL-Context and NYUD, and also demonstrates superior incremental learning capability as compared to its close competitors. The code and models are made publicly available\footnote{\url{https://github.com/menelaoskanakis/RCM}}.

\keywords{Multi-Task Learning, Incremental Learning, Task Interference}
\end{abstract}

%% file: text/intro.tex
\section{Introduction}
\label{sec:intro}

\begin{figure*}[h!]
 \centering
   \subfloat[Single-Task setup]{\label{fig:STS}
      \includegraphics[width=.32\textwidth]{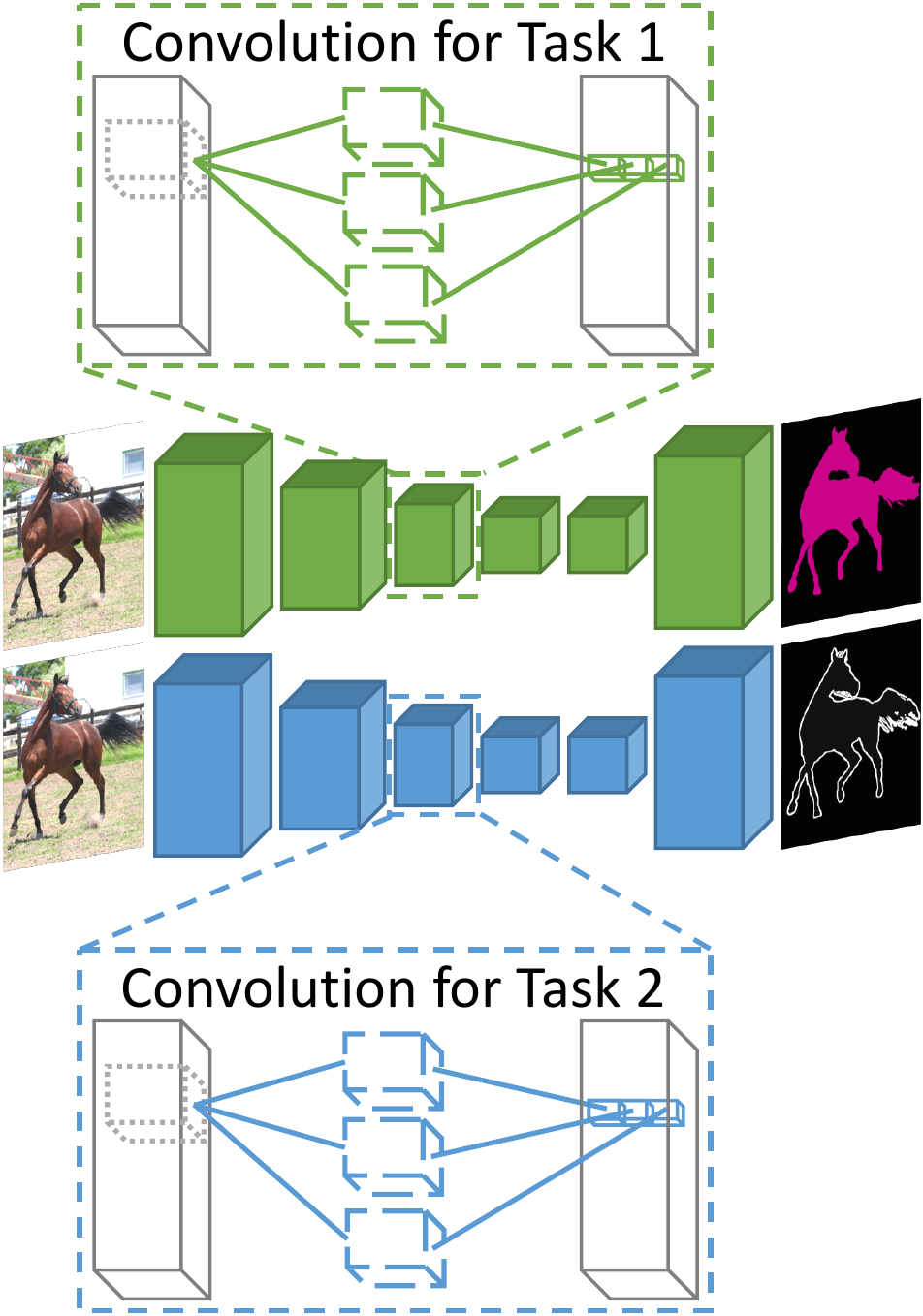}}
~
   \subfloat[Multi-Task setup]{\label{fig:MTS}
      \includegraphics[width=.32\textwidth]{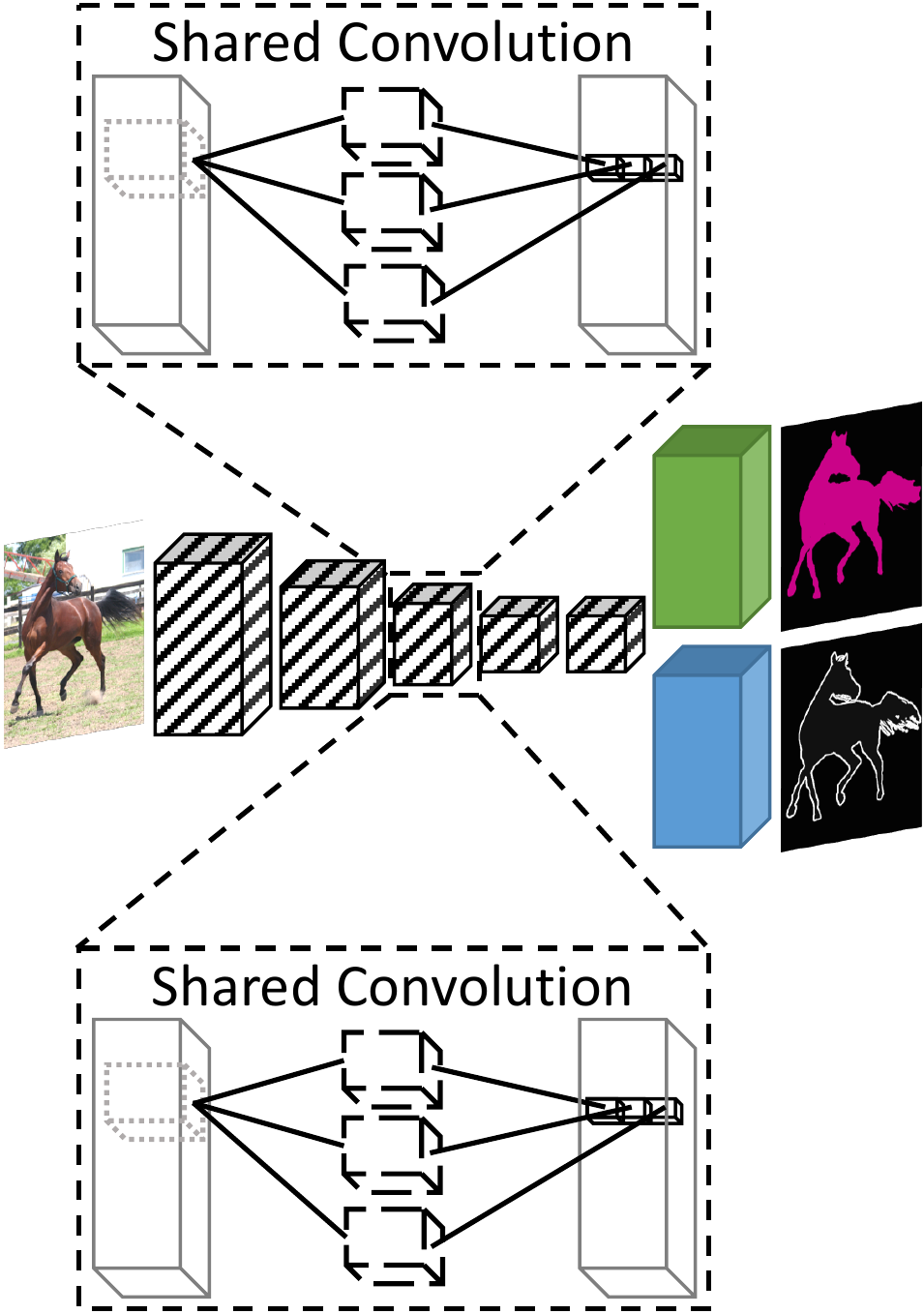}}
~
       \subfloat[RCM setup (ours)]{\label{fig:RCM_ours}
      \includegraphics[width=.32\textwidth]{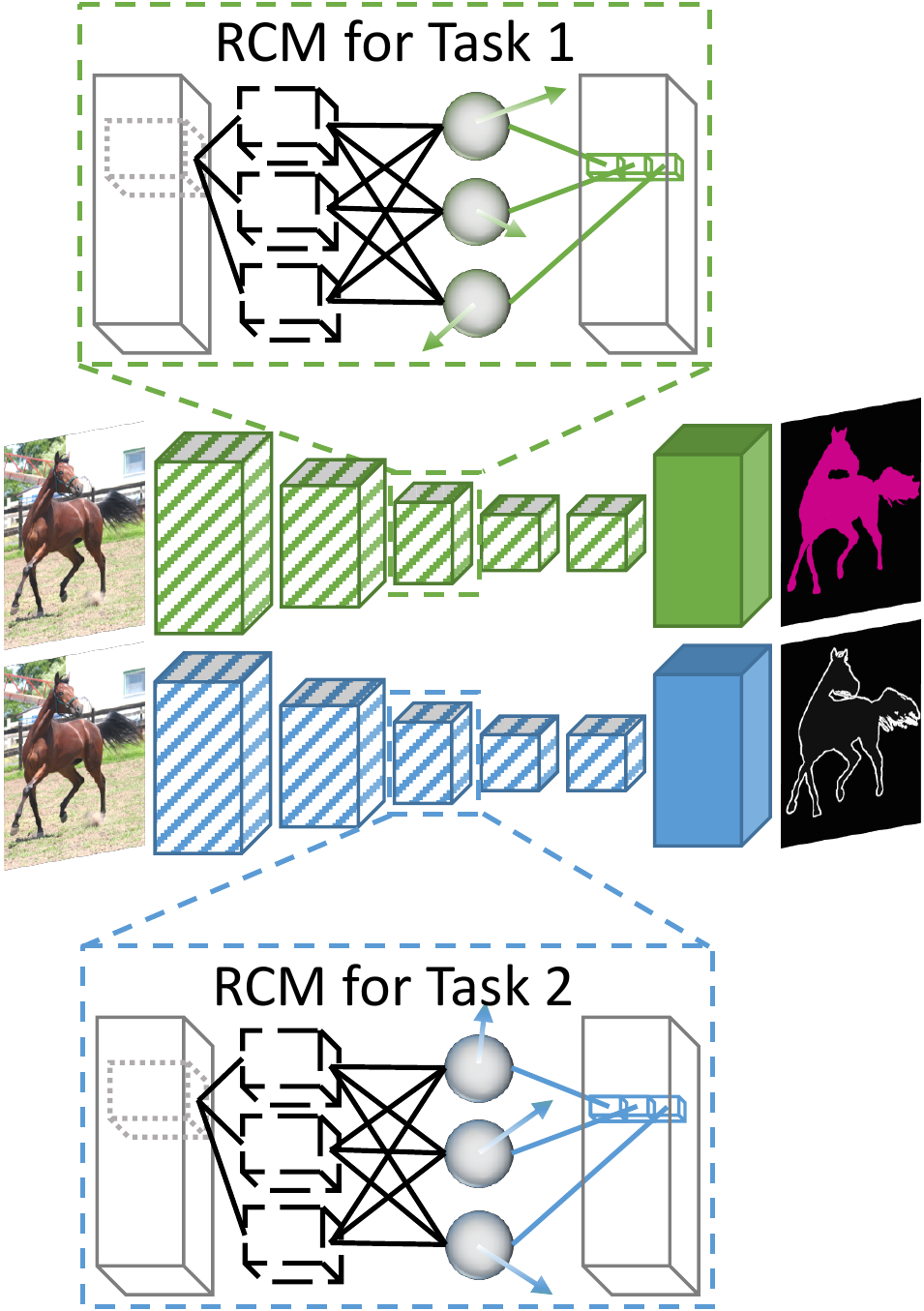}}

\caption{\textbf{(a)} Optimizing independent models per task allows for the easy addition of new tasks, at the expense of a multiplicative increase in the total number of parameters with respect to a single model (green and blue denote task-specific parameters). \textbf{(b)} A single backbone for multiple tasks must be meaningful to all, thus, all tasks interact with the said backbone (black indicates common parameters). \textbf{(c)} Our proposed setup, RCM (Reparameterized Convolutions for Multi-task learning), uses a pre-trained filter bank (denoted in black) and independently optimized task-specific modulators(denoted in colour) to adapt the filter bank on a per-task basis. New task addition is accomplished by training the task-specific modulators, thus explicitly addressing task interference while parameters scale at a slower rate than having independent models per task.}\label{fig:overview}
   
\end{figure*}

Over the last decade, convolutional neural networks (CNNs) have been established as the standard approach for many computer vision tasks, like image classification~\cite{krizhevsky2012imagenet,simonyan2015very,he2016deep}, object detection~\cite{girshick2014rich,redmon2016you,liu2016ssd}, semantic segmentation~\cite{long2015fully,chen2017deeplab,zhao2017pyramid}, and monocular depth estimation~\cite{eigen2014depth,laina2016deeper}. Typically, these tasks are handled by CNNs independently, i.e., a separate model is optimized for each task, resulting in several task-specific models (Fig.~\ref{fig:STS}). However, real-world problems are more complex and require models to perform multiple tasks on-demand without significantly compromising each task's performance. For example, an interactive advertisement system tasked with displaying targeted content to its audience should be able to detect the presence of humans in its viewpoint effectively, estimate their gender and age group, recognize their head pose, etc. At the same time, there is a need for flexible models able to gradually add more tasks to their knowledge, without forgetting previously known tasks or having to re-train the whole model from scratch. For instance, a car originally deployed with lane and pedestrian detection functionalities can be extended with depth estimation capabilities post-production.

When it comes to learning multiple tasks under a single model, multi-task learning (MTL) techniques~\cite{caruana1997multitask,ruder2017overview} have been employed in the literature. On the one hand, encoder-focused approaches~\cite{misra2016cross,kokkinos2017ubernet,lu2017fully,doersch2017multi,neven2017fast,liu2019end,bragman2019stochastic,vandenhende2019branched} emphasize learning feature representations from multi-task supervisory signals by employing architectures that encode shared and task-specific information. On the other hand, decoder-focused approaches~\cite{xu2018pad,zhang2018joint,zhang2019pattern,vandenhende2020mti} utilize the multi-task feature representations learned at the encoding stage to distill cross-task information at the decoding stage, thus refining the original feature representations. In both cases, however, the joint learning from multiple supervisory signals (i.e., tasks) can hinder the individual task performance if the associated tasks point to conflicting gradient directions during the update step of the shared feature representations (Fig.~\ref{fig:MTS}). Formally this is known as \textit{task interference} or \textit{negative transfer} and has been well documented in the literature~\cite{kokkinos2017ubernet,maninis2019attentive,zhao2018modulation}. To suppress negative transfer, several approaches~\cite{chen2017gradnorm,kendall2018multi,sinha2018gradient,guo2018dynamic,zhao2018modulation,sener2018multi,maninis2019attentive} dynamically re-weight each task's loss function or re-order the task learning, to find a `sweet spot' where individual task performance does not degrade significantly. Arguably, such approaches mainly focus on mitigating the negative transfer problem in the MTL architectures above, rather than eliminating it (see Section~\ref{sec:task_interf}). At the same time, existing works seem to disregard the fact that MTL models are commonly desired to be incremental, i.e., information from new tasks should be continuously incorporated while existing task knowledge is preserved. In existing works, the MTL model has to be re-trained from scratch if the task dictionary changes; this is arguably sub-optimal. 

Recently, task-conditional networks~\cite{maninis2019attentive} emerged as an alternative for MTL, inspired by work in multi-domain learning~\cite{rebuffi2017learningRA1,rebuffi2018efficientRA2}. That is, performing separate forward passes within an MTL model, one for each task, every time activating a set of task-specific residual responses on top of the shared responses. Note that, this is useful for many real-world setups (e.g., an MTL model deployed in a mobile phone with limited resources that adapts its responses according to the task at hand), and particularly for incremental learning (e.g., a scenario where the low-level tasks should be learned before the high-level ones). However, the proposed architecture in~\cite{maninis2019attentive} is prone to task interference due to the inherent presence of shared modules, which is why the authors introduced an adversarial learning scheme on the gradients to minimize the performance degradation. Moreover, the model needs to be trained from scratch if the task dictionary changes.

All given, existing works primarily focus on either improving the multi-task performance or reducing the number of parameters and computations in the MTL model. In this paper, we take a different route and explicitly tackle the problems of incremental learning and task interference in MTL. We show that both problems can be addressed simply by reparameterizing the convolutional operations of a neural network. In particular, building upon the task-conditional MTL direction, we propose to decompose each convolution into a shared part that acts as a filter bank encoding common knowledge, and task-specific modulators that adapt this common knowledge uniquely for each task. Fig.~\ref{fig:RCM_ours} illustrates our approach, RCM (Reparameterized Convolutions for Multi-task learning). Unlike existing works, the shared part in our case is not trainable to explicitly avoid negative transfer. Most notably, as any number of task-specific modulators can be introduced in each convolution, our model can incrementally solve more tasks without interfering with the previously learned ones. Our results demonstrate that the proposed RCM can outperform state-of-the-art methods in multi-task (Section~\ref{subsec:SOTA}) and incremental learning (Section~\ref{subsec:IL}) experiments. At the same time, we address the common multi-task challenge of task interference by construction, by ensuring tasks can only update task-specific components and cannot interact with each other.

%% file: text/related_work.tex
\section{Related Work}
\label{sec:rel_work}

\noindent
\textbf{Multi-task learning (MTL)} aims at developing models that can solve a multitude of tasks~\cite{caruana1997multitask,ruder2017overview}. In computer vision, MTL approaches can roughly be divided into encoder-focused and decoder-focused ones. Encoder-focused approaches primarily emphasize on architectures that can encode multi-purpose feature representations through supervision from multiple tasks. Such encoding is typically achieved, for example, via feature fusion~\cite{misra2016cross}, branching~\cite{kokkinos2017ubernet,neven2017fast,lu2017fully,vandenhende2019branched}, self-supervision~\cite{doersch2017multi}, attention~\cite{liu2019end}, or filter grouping~\cite{bragman2019stochastic}. Decoder-focused approaches start from the feature representations learned at the encoding stage, and further refine them at the decoding stage by distilling information across tasks in a one-off~\cite{xu2018pad}, sequential~\cite{zhang2018joint}, recursive~\cite{zhang2019pattern}, or even multi-scale~\cite{vandenhende2020mti} manner. Due to the inherent layer sharing, the approaches above typically suffer from task interference. Several works proposed to dynamically re-weight the loss function of each task~\cite{chen2017gradnorm,kendall2018multi,sinha2018gradient,sener2018multi}, sort the order of task learning~\cite{guo2018dynamic}, or adapt the feature sharing between `related' and `unrelated' tasks~\cite{zhao2018modulation}, to mitigate the effect of negative transfer. In general, existing MTL approaches have primarily focused on improving multi-task performance or reducing the network parameters and computations. Instead, in this paper, we look at the largely unexplored problems of incremental learning and negative transfer in MTL models and propose a principled way to tackle them. \\

\noindent
\textbf{Incremental learning (IL)} is a paradigm that attempts to augment the existing knowledge by learning from new data. IL is often used, for example, when aiming to add new classes~\cite{rebuffi2017icarl} to an existing model, or learn new domains~\cite{li2017learning}. It aims to mitigate `catastrophic forgetting'~\cite{french1999catastrophic}, the phenomenon of forgetting old tasks as new ones are learned. To minimize the loss of existing knowledge, Li and Hoiem~\cite{li2017learning} optimized the new task while preserving the old task's responses. Other works~\cite{kirkpatrick2017overcoming,lee2017overcoming} constrained the optimization process to minimize the effect learning has on weights important for older tasks. Rebuffi et al.~\cite{rebuffi2017icarl} utilized exemplars that best approximate the mean of the learned classes in the feature space to preserve performance. Note that the performance of such techniques is commonly upper bounded by the joint training of all tasks. More relevant to our work, in a multi-domain setting, a few approaches~\cite{rebuffi2017learningRA1,rebuffi2018efficientRA2,rosenfeld2018incremental,mallya2018piggyback} utilize a pre-trained network that remains untouched and instead learn domain-specific components that adapt the behavior of the network to address the performance drop common in IL techniques. Inspired by this research direction, we investigate the training of parts of the network, while keeping the remaining components constant from initialization amongst all tasks. This technique not only addresses catastrophic forgetting but also task interference, which is crucial in MTL. \\

\noindent
\textbf{Decomposition} of filters and tensors within CNNs has been explored in the literature. In particular, filter-wise decomposition into a product of low-rank filters~\cite{jaderberg2014speeding}, filter groups~\cite{peng2018extreme}, a basis of filter groups~\cite{li2019learning}, etc.~have been utilized. In contrast, tensor-wise examples include SVD decomposition~\cite{denton2014exploiting,zhang2015accelerating}, CP-decomposition~\cite{lebedev2014speeding}, Tucker decomposition~\cite{kim2015compression}, Tensor-Train decomposition~\cite{oseledets2011tensor}, Tensor-Ring decomposition~\cite{zhao2016tensor}, T-Basis~\cite{obukhov2020tbasis}, etc. These techniques have been successfully used for compressing neural networks or reducing their inference time. Instead, in this paper, we utilize decomposition differently. We decompose each convolutional operation into two components: a shared and a task-specific part. Note that although we utilize the SVD decomposition for simplicity, the same principles hold for other decomposition types too. 

%% file: text/methodology.tex
\section{Reparameterizing CNNs for Multi-Task Learning}
\label{sec:method}

In this section, we present techniques to adapt a CNN architecture, such that it can increasingly learn new tasks in an MTL setting while scaling more efficiently than simply adding single-task models. Section~\ref{sec:prob_form} introduces the problem formulation. Section~\ref{sec:task_interf} demonstrates the effect of task interference in MTL and motivates the importance of CNN reparameterization. Section~\ref{sec:reparam-cov} presents techniques to reparameterize CNNs and limit the parameter increase with respect to task-specific models.

\begin{figure*}[t]
 \centering
  \subfloat[Standard Conv.]{\label{fig:convST}
      \includegraphics[width=.31\textwidth]{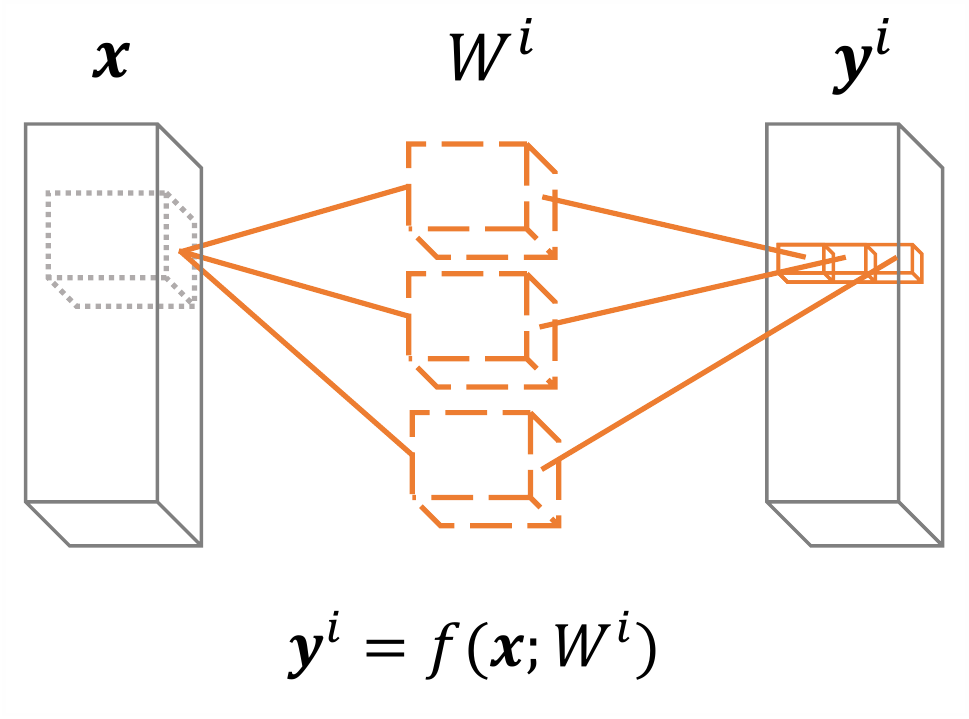}}
~
  \subfloat[RC without NFF]{\label{fig:RC}
      \includegraphics[width=.31\textwidth]{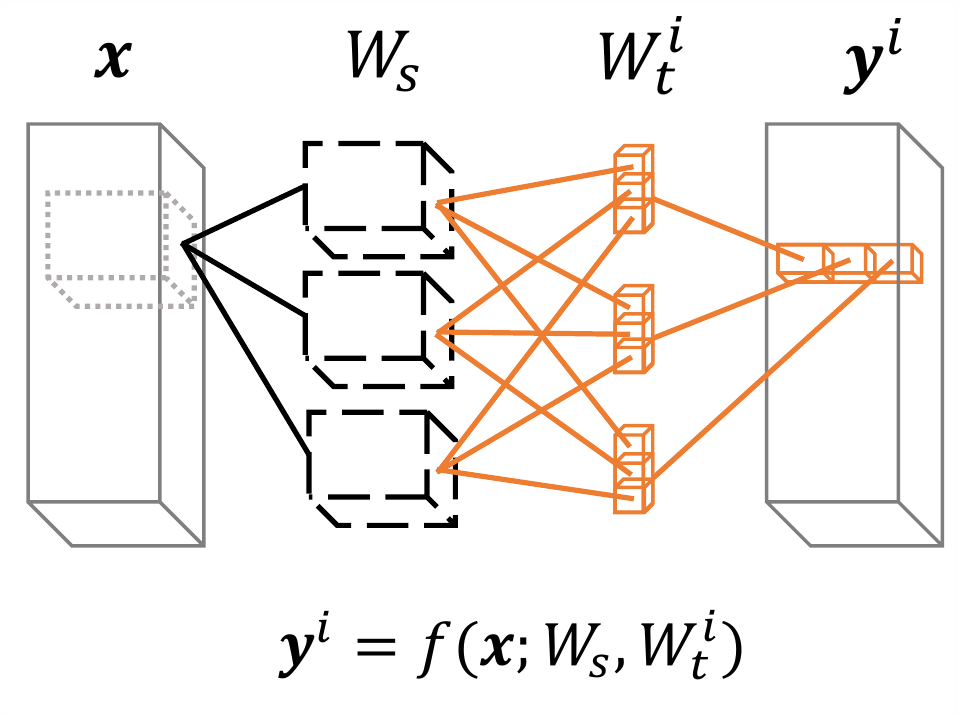}}
~
      \subfloat[RC with NFF]{\label{fig:RC_NFF}
      \includegraphics[width=.31\textwidth]{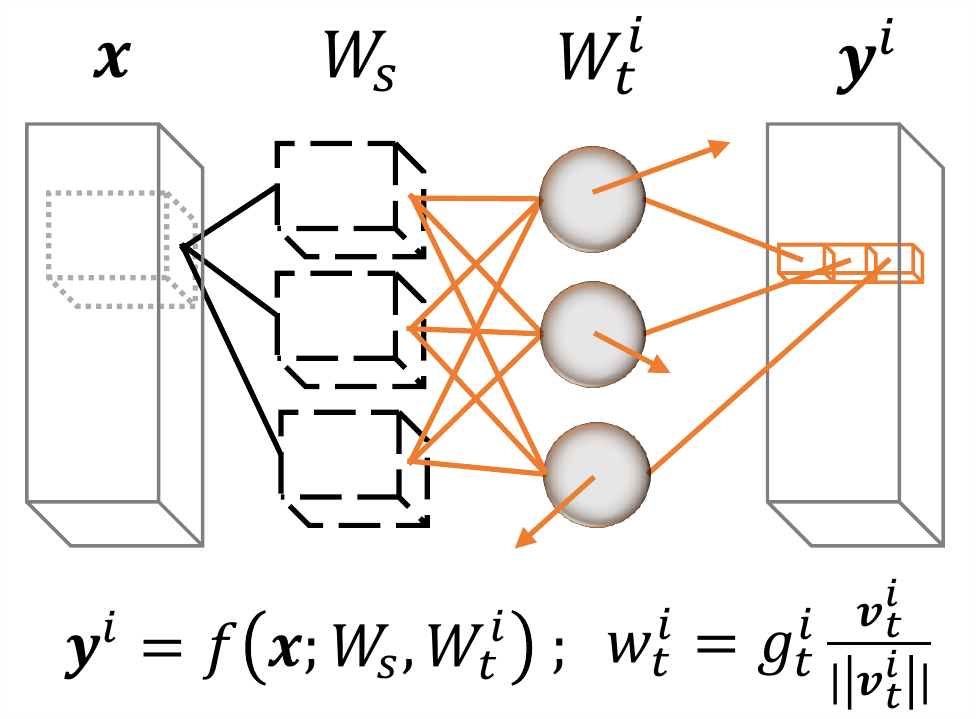}}

\caption{\textbf{(a)} A standard convolutional module for a given task i, with task-specific weights $W^i$ in orange. \textbf{(b)} A reparameterized convolution (RC) consisting of a shared filter bank $W_s$ in black, and task-specific modulator $W_t^i$ in orange. \textbf{(c)} An RC with Normalized Feature Fusion (NFF), consisting of a shared filter bank $W_s$ in black, and task-specific modulator $W_t^i$ in orange. Each row $\boldsymbol{w}_t^i$ of $W_t^i$ is reparameterized as $g_{t}^i \cdot \boldsymbol{v_{t}^i}/\parallel\boldsymbol{v_{t}^i}\parallel$.}
   
\end{figure*}

\subsection{Problem Formulation}
\label{sec:prob_form}
Given $P$ tasks and input tensor $\boldsymbol{x}$, we aim to learn a function $f(\boldsymbol{x}; W_{s}, W_{t}^{i}) = \boldsymbol{y}^{i}$ that holds for task $i = 1, 2, \dots P$, where $W_{s}$ and $W_{t}^{i}$ are the shared and task-specific parameters respectively. Unlike existing approaches~\cite{lu2017fully,misra2016cross} which learn such functions $f(\cdot)$ on the layer level of the network, i.e., explicitly designing shared and task-specific layers, we aim to learn $f$ on a block-level by \emph{reparameterizing} the convolutional operation, and adapting its behaviour conditioned on the task $i$, as depicted in Fig.~\ref{fig:RC} and Fig.~\ref{fig:RC_NFF}. By doing so, we can explicitly address the task interference and catastrophic forgetting problems within an MTL setting.

\subsection{Task Interference}
\label{sec:task_interf}
To motivate the importance of addressing task interference by construction, we analyze the task-specific gradient directions on the shared modules of a state-of-the-art MTL model. Specifically, we utilize the work of \cite{maninis2019attentive}, who used a discriminator to enforce indistinguishable gradients amongst tasks.

\begin{figure}[t]
\centering
    \includegraphics[width=\textwidth]{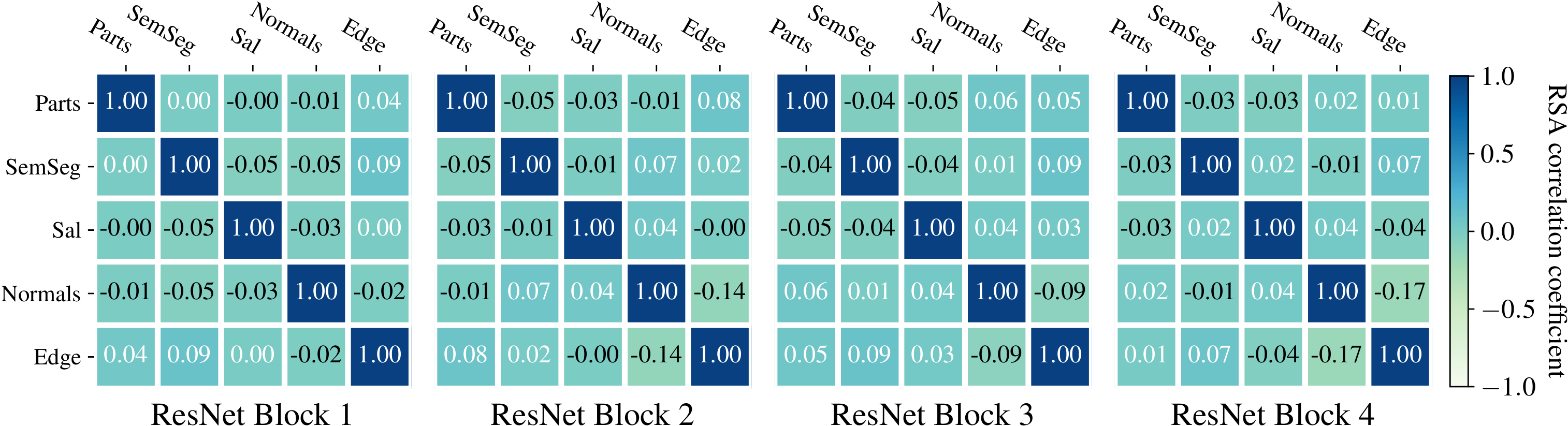}
\caption{Visualization of the Representation Similarity Analysis (RSA) on the task-specific gradients at different depths of a ResNet-26 model~\cite{maninis2019attentive}. The analysis was conducted on: human parts segmentation (Parts), semantic segmentation (SemSeg), saliency estimation (Sal), normals estimation (Normals), and edge detection (Edge).}
\label{fig:gradients}
\end{figure}

We acquire the gradients from the training dataset of PASCAL-Context~\cite{mottaghi2014role} for each task, using minibatches of size 128, yielding 40 minibatches. We then use the Representation Similarity Analysis (RSA), proposed in~\cite{dwivedi2019representation} for transfer learning, as a means to quantify the correlation of the gradients amongst the different tasks. Fig.~\ref{fig:gradients} depicts the task gradient correlations at different depths of a ResNet-26 model \cite{he2016deep}, trained to have indistinguishable gradients in the output layer~\cite{maninis2019attentive}. It can be seen that there is a limited gradient correlation amongst the tasks, demonstrating that addressing task interference indirectly (here with the use of adversarial learning on the gradients) is a very challenging problem. We instead follow a different direction and propose to utilize reparameterizations with shared components amongst different tasks that are untouched during the training process, and each task being able to optimize only its parameters. As such, task interference is eliminated by construction.

\subsection{Reparameterizing Convolutions} \label{sec:reparam-cov}

We define a convolutional operation $f(\boldsymbol{x}; \boldsymbol{w})=y$ for the single-task learning setup, Fig.~\ref{fig:convST}. $\boldsymbol{w} \in \mathbb{R}^{k^2 c_{in}}$ denotes the parameters of a single convolutional layer (we omit the bias to simplify notation) for a kernel size $k$
and $c_{in}$ channels. $\boldsymbol{x} \in \mathbb{R}^{k^2 c_{in}}$ is the input tensor volume at a given spatial location ($\boldsymbol{x}$ and $\boldsymbol{w}$ are expressed in vector notation), and $y$ is the scalar response. 
Assuming $c_{out}$ such filters, the convolutional operator can be rewritten in matrix notation as $f(\boldsymbol{x}; W)=\boldsymbol{y}$, where $\boldsymbol{y}\in \mathbb{R}^{c_{out}}$ provides $c_{out}$ responses, and $W \in \mathbb{R}^{c_{out} \times k^2c_{in}}$.
In a single-task setup:
\begin{align}
f(\boldsymbol{x}; W^{1})=\boldsymbol{y}^{1},~\dots~,~f(\boldsymbol{x}; W^{P})=\boldsymbol{y}^{P} \label{eq:mtl1}
\end{align}
where $W^{i}$ and $\boldsymbol{y}^{i}$ are the task-specific parameters and responses for a given convolutional layer, respectively. 
The total number of parameters for this setup is  $\mathcal{O}(Pk^2c_{in}c_{out})$.
Our goal is to reparameterize $f(\cdot)$ in Eqn.~\ref{eq:mtl1} as:
\begin{align}
f(\boldsymbol{x}; W^{i})= h(\boldsymbol{x}; W_{s}, W_{t}^{i}), \quad \forall i = 1, \dots, P \label{eq:mtl2}
\end{align}
using a set of shared ($W_{s} \in \mathbb{R}^{c_{out} \times k^{2}c_{in}}$) and task-specific ($W_{t}^{i} \in \mathbb{R}^{c_{out} \times c_{out}}$) parameters for each convolutional layer of the backbone. Our formulation aims to retain the prediction performance of the original convolutional layer (Eq.~\ref{eq:mtl1}), while simultaneously reducing the rate in which the total number of parameters grows. The complexity now becomes $\mathcal{O}((k^2c_{in} + Pc_{out})c_{out})$, which is less than $\mathcal{O}(Pk^2c_{in}c_{out})$ for standard layers.
We argue that this reparameterization is necessary for coping with task interference and incremental learning in an MTL setup, in which we only optimize for task-specific parameters $W_{t}^{i}$, while keeping the shared parameters $W_{s}$ intact. 
Note that, when adding a new task $i=\omega$, we do not need to train the entire network from scratch as in~\cite{maninis2019attentive}. We only
optimize $W_{t}^{\omega}$ for each layer of the reparameterized CNN.

We denote our reparameterized convolutional layer as a matrix multiplication between the two sets of parameters: $W_{t}^{i} W_{s}$. In order to find a set of parameters $W_{t}^{i} W_{s}$ that approximates the single-task weights $W^{i}$ a natural choice is to minimize the Frobenius norm $\parallel$$W_{t}^{i} W_{s}-W^{i}$$\parallel_F$ directly.
Even though direct minimization of this metric is appealing due to its simplicity, it poses some major caveats.
(i) It assumes that all directions in the parameter space affect the final performance for task $i$ in the same way and are thus penalized uniformly. However, two different solutions for $W_{t}^{i}$ with the same Frobenius norm can yield drastically different losses.
(ii) This approximation is performed independently for each convolutional layer, neglecting the chain effect an inaccurate prediction in one layer can have in the succeeding layers. In the remainder of this section, we propose different techniques to address these limitations.\\

\noindent 
\textbf{Reparameterized Convolution.}
We implement the Reparameterized Convolution (RC) $W_{t}^{i} W_{s}$ as a stack of two 2D convolutional layers without non-linearity in between, with $W_{s}$ having a spatial filter size $k$ and $W_{t}^{i}$ being a $1 \times 1$ convolution (Fig.~\ref{fig:RC})\footnote{To ensure compliance with ImageNet~\cite{deng2009imagenet} initialization, the new architecture is first pre-trained on ImageNet using the publicly available training script from PyTorch~\cite{paszke2019pytorch}.}. We optimize only $W_{t}^{i}$ directly on the task-specific loss function using stochastic gradient descent while keeping the shared weights $W_{s}$ constant. This ensures that training for one task is independent of other tasks, ruling out interference amongst tasks while optimizing the metric of interest.\\

\noindent 
\textbf{Normalized Feature Fusion.} One can view $\boldsymbol{w}_{t}^{i}$, a row in matrix $W_{t}^{i}$, as a soft filter adaptation mechanism, i.e., a modulator which generates new task-specific filters from a given filter bank $W_{s}$, depicted in Fig.~\ref{fig:RC}. However, instead of training the vector $\boldsymbol{w}_{t}^{i}$ directly, we propose its reparameterization into two terms, a vector term $\boldsymbol{v}_{t}^{i}\in \mathbb{R}^{c_{out}}$, and a scalar term $g_{t}^{i}$ as:
\noindent 
\begin{align}
	\label{eq:weight_norm}
	\boldsymbol{w}_{t}^{i} = g_{t}^{i} \dfrac{\boldsymbol{v}_{t}^{i}}{\parallel \boldsymbol{v}_{t}^{i} \parallel}, 
\end{align}
where $\parallel\cdot\parallel$ denotes the Euclidean norm. We refer to this reparameterization as Normalized Feature Fusion (NFF), depicted in Fig.~\ref{fig:RC_NFF}. NFF provides an easier optimization process in comparison to an unconstrained $\boldsymbol{w}_{t}^{i}$. This reparametrization enforces ${\boldsymbol{v}_{t}^{i}}/{\parallel \boldsymbol{v}_{t}^{i} \parallel}$ to be unit length and point in the direction which best merges the filter bank. The vector norm $\parallel \boldsymbol{w}_{t}^{i} \parallel=$ $g_{t}^{i}$ learns independently the appropriate scale of the newly generated filters, and thus the scale of the activation. Directly optimizing $\boldsymbol{w}_{t}^{i}$ attempts to learn both jointly, which is a harder optimization problem. Normalizing weight tensors has been generally explored for speeding up the convergence of the optimization process~\cite{dauphin2017language,salimans2016weight,srebro2005rank}. In our work, we use it differently and demonstrate empirically (see Section~\ref{subsec:Abl_RCM}) that such a reparameterization in series with a filter bank also improves performance in the MTL setting. As seen in Eq.~\ref{eq:weight_norm}, additional learnable parameters are introduced in the training process ($g_{t}^{i}$ and $\boldsymbol{v}_{t}^{i}$), however, $\boldsymbol{w}_{t}^{i}$ can be computed after training and used directly for deployment, eliminating additional overhead.\\

\noindent 
\textbf{Response Initialization.}
We build upon the findings of matrix/tensor decomposition literature~\cite{denton2014exploiting,zhang2015accelerating} that network weights/responses lie on a low dimensional subspace. We further assume that such a subspace can be beneficial for multiple tasks, and thus good for network initialization under a MTL setup. To this end, we identify a meaningful subspace of the responses for the generation of a better filter bank $W_{s}$ when compared to that directly learned by pre-training $W_{s}$ on ImageNet. More formally, let $\boldsymbol{y} = f(\boldsymbol{x}; W^{m})$ be the responses for input tensor $\boldsymbol{x}$, where $W^{m} \in \mathbb{R}^{c_{out} \times k^{2}c_{in}}$ are the pre-trained ImageNet weights. We define $Y \in \mathbb{R}^{c_{out} \times n}$ as a matrix containing $n$ responses of $\boldsymbol{y}$ with the mean vector $\overline{\boldsymbol{y}}$ subtracted. We compute the  eigen-decomposition of the covariance matrix $YY^{T} = USU^{T}$ (using Singular Value Decomposition, SVD), where $U \in \mathbb{R}^{c_{out} \times c_{out}}$ is an orthogonal matrix with the eigenvectors on the columns, and $S$ is a diagonal matrix of the corresponding eigenvalues. We can now initialize the shared convolution parameters $W_{s}$ with $U^{T}W^{m}$, and the task-specific $W_{t}^{i}$ with $U$. We refer to this initialization methodology as Response Initialization (RI). We point the reader to the supplementary material for more details.

%% file: text/exp.tex
\section{Experiments}
\label{sec:experiments}

\subsection{Datasets} We focus our evaluation on dense prediction tasks, making use of two datasets. We conduct the majority of the experiments on PASCAL~\cite{everingham2010pascal}, and more specifically, PASCAL-Context~\cite{mottaghi2014role}. We address edge detection (Edge), semantic segmentation (SemSeg), human parts segmentation (Parts), surface normals estimation (Normals), and saliency (Sal). We evaluate single-task performance using optimal dataset F-measure (odsF)~\cite{martin2004learning} for edge detection, mean intersection over union (mIoU) for semantic segmentation, human parts and saliency, and finally mean error (mErr) for surface normals. Labels for human parts segmentation are acquired from~\cite{chen2014detect}, while for saliency and surface normals from~\cite{maninis2019attentive}.

We further evaluate the proposed method on the smaller NYUD dataset~\cite{silberman2012indoor}, comprised of indoor scenes, on edge detection (Edge), semantic segmentation (SemSeg), surface normals estimation (Normals), and depth (Depth). The evaluation metrics for edge detection, semantic segmentation, and surface normals estimation are identical to those for PASCAL-Context, while for depth we use root mean squared error (RMSE). 

\subsection{Architecture} All of our experiments make use of the DeepLabv3+ architecture~\cite{chen2018encoder}, originally designed for semantic segmentation, which performs competitively for all tasks of interest as demonstrated in~\cite{maninis2019attentive}. DeepLabv3+ encodes multi-scale contextual information by utilizing a ResNet~\cite{he2016deep} encoder with a-trous convolutions~\cite{chen2017deeplab} and an a-trous spatial pyramid pooling (ASPP) module, while a decoder with a skip connection refines the predictions. Unless otherwise stated, we use a ResNet-18 (R-18) based DeepLabv3+, and report the mean performance of five runs for each experiment\footnote{Baseline comparisons to competing methods, as well as additional backbone experiments, can be found in the supplementary material.}.

\noindent
\subsection{Evaluation Metric} We follow standard practice~\cite{maninis2019attentive,vandenhende2020mti} and quantify the performance of a model $m$ as the average per-task performance drop with respect to the corresponding single-task baseline $b$:
\begin{align}
\label{eq:metric}
\Delta _m= \frac{1}{P}\sum_{i=1}^{P}(-1)^{l_i} \frac{M_{m,i}-M_{b,i}}{M_{b,i}}
\end{align}

\noindent
where $l_i$ is either 1 or 0 if a lower or a greater value indicates better performance, respectively, for a performance measure $M$. P indicates the total number of tasks.

\subsection{Analysis of network module sharing} 
\label{subsection:ModuleAnalysis}
\begin{wrapfigure}{R}{0.5\textwidth}
\centering
\includegraphics[width=0.5\textwidth]{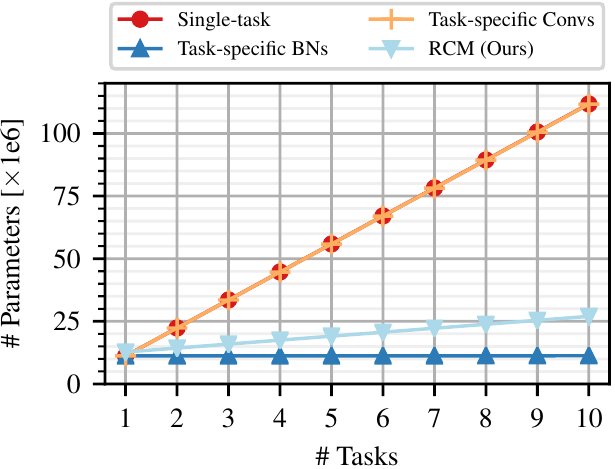}
\caption{\textbf{Backbone parameter scaling.} Total number of parameters with respect to the number of tasks for R-18 backbone.}
\label{fig:params_task}
\end{wrapfigure}

We investigate the level of task-specific adaptation required for a common backbone to perform competitively to single-task models, while additionally eliminating negative transfer. In other words, the necessity for task-specific modules, i.e., convolutions (Convs) and batch normalizations (BNs) \cite{ioffe2015batch}. Specifically, we optimize for task-specific Convs, BNs, or both along the network's depth. Modules that are not being optimized maintain their ImageNet pre-trained parameters. Table~\ref{table:ablation_norm_conv} presents the effect on performance, while Fig.~\ref{fig:params_task} depicts the total number of parameters with respect to the number of tasks. Experiments vary from common Convs and BNs (Freeze encoder) to task-specific Convs and BNs (Single-task), and anything in-between.

The model utilizing a common backbone pre-trained on ImageNet (Freeze encoder), as expected, is unable to perform competitively to the single-task counterpart, with a performance drop of 14.98\%. Task-specific BNs significantly improve performance with a percentage drop of 5.76\%, at a minimal increase in parameters (Fig.~\ref{fig:params_task}). The optimization of Convs is essential for competitive performance to single-task, with a percentage drop of 0.62\%. However, the increase in parameters is comparable to single-task, which is undesirable (Fig.~\ref{fig:params_task}).

\setlength{\tabcolsep}{4pt}
\begin{table}[t]
\begin{center}
\caption{\textbf{Performance analysis of task-specific modules.} We report the effect network modules (Convs and BNs) have on the performance of PASCAL-Context.}
\label{table:ablation_norm_conv}
\scalebox{0.9}{
\begin{tabular}{l|cc||ccccc|c}
 Method & Convs & BNs & Edge $\uparrow$ & SemSeg $\uparrow$ & Parts $\uparrow$ & Normals $\downarrow$ & Sal $\uparrow$ & $\Delta _m \%$ $\downarrow$ \\
\hline
Freeze encoder &  &  & 67.32 & 60.37 & 47.86 & 17.40 & 58.39 & 14.98 \\
Task-specific BNs &  & \checkmark & 69.80 & 63.93 & 53.22 & 14.78 & 64.44 & 5.76 \\
Task-specific Convs & \checkmark &  & 71.72 & 66.00 & 59.05 & 13.78 & 66.31 & 0.62 \\
Single-task & \checkmark & \checkmark & 71.88 & 66.22 & 59.69 & 13.64 & 66.62 & - \\
\end{tabular}
}
\end{center}
\end{table}
\setlength{\tabcolsep}{1.4pt}

\subsection{Ablation study} 
\label{subsec:Abl_RCM}
To validate the proposed methodology from Section~\ref{sec:method}, we conduct an ablation study, presented in Table~\ref{table:adaptation}. We additionally report the performance of a model trained jointly on all tasks, consisting of a fully shared encoder and task-specific decoders (Multi-task). This multi-task model is not trained in an IL setup but merely serves as a reference to the traditional multi-tasking techniques. We report a performance drop of 3.32\% with respect to the single-task setup. \\
\setlength{\tabcolsep}{4pt}
\begin{table}[t]
\begin{center}
\caption{\textbf{Ablation study of the proposed RCM.} 
We present ablation experiments for the proposed  Reparameterized Convolution (RC), Response Initialization (RI), Normalized Feature Fusion (NFF) on PASCAL-Context dataset.}
\label{table:adaptation}
\scalebox{0.9}{
\begin{tabular}{l|cc||ccccc|c}
 Method & NFF & RI & Edge $\uparrow$ & SemSeg $\uparrow$ & Parts $\uparrow$ & Normals $\downarrow$ & Sal $\uparrow$ & $\Delta _m \%$ $\downarrow$ \\
\hline
Single-task & & & 71.88 & 66.22 & 59.69 & 13.64 & 66.62 & - \\
Multi-task & & & 70.74 & 62.43 & 57.89 & 14.43 & 66.31 & 3.32 \\
\hline
RC & & & 71.10 & 64.56 & 56.87 & 13.91 & 66.37 & 2.13 \\
RC+NFF & \checkmark & & 71.12 & 64.71 & 56.91 & 13.90 & 66.33 & 2.07 \\
RC+RI & & \checkmark & 71.36 & 65.58 & 57.99 & 13.70 & 66.21 & 1.12 \\
RC+RI+NFF & \checkmark & \checkmark & 71.34 & 65.70 & 58.12 & 13.70 & 66.38 & 0.99 \\
\end{tabular}
}
\end{center}
\end{table}
\setlength{\tabcolsep}{1.4pt}

\noindent
\textbf{Reparameterized Convolution.} We first develop a new baseline for our proposed reparameterization, where we replace every convolution with the RC (Section~\ref{sec:reparam-cov}) counterpart. As seen in Table~\ref{table:adaptation}, RC achieves a performance drop of 2.13\%, outperforming the 3.32\% drop of the multi-task baseline, as well as the Task-specific BNs (Table~\ref{table:ablation_norm_conv}) that achieved a performance drop of 5.76\%. This observation corroborates the claim made in Section~\ref{subsection:ModuleAnalysis} that task-specific adaptation of the convolutions is essential for a model to perform competitively for all tasks. Additionally, we demonstrate that even without training entirely task-specific convolutions, as in Table~\ref{table:ablation_norm_conv} (Task-specific Convs), a performance boost can still be observed at a smaller magnitude, while the total number of parameters scales at a slower rate (Fig.~\ref{fig:params_task}). RCM in Fig.~\ref{fig:params_task} depicts the parameter scaling of all the RC-based methods introduced in Table~\ref{table:adaptation}, described in this section. As such, improvements in performance from this baseline do not stem from an increase in network capacity.\\

\noindent
\textbf{Response Initialization.} We investigate the effect on the performance of a more meaningful filter bank, RI (Section~\ref{sec:reparam-cov}), against the filter bank learned by directly pre-training the RC architecture on ImageNet. In Table~\ref{table:adaptation} we report the performance of our proposed model when directly pre-trained on ImageNet (Table~\ref{table:adaptation}-RC), and with the RI based filter bank (Table~\ref{table:adaptation}-RC+RI). Compared to the RC model, the performance significantly improves from a 2.13\% drop to a 1.12\% drop with the RC+RI model. This observation clearly demonstrates that the filter bank generated using our proposed RI approach is beneficial for better weight initialization.\\

\noindent
\textbf{Normalized Feature Fusion.} We replace the unconstrained task-specific components of RC with the proposed NFF (Section~\ref{sec:reparam-cov}). We demonstrate in Table~\ref{table:adaptation} that NFF improves the performance no matter the initialization of the filter bank. RC improves from a 2.13\% drop to a 2.07\% in RC+NFF, while RC+RI improved from a 1.12\% drop to 0.99\% for RC+RI+NFF.

The architecture used for the remaining experiments is the Reparameterized Convolution (RC) with Normalized Feature Fusion (NFF), initialized using the Response Initialization (RI) methodology. This architecture is denoted as RCM.

\subsection{Comparison to state-of-the-art}
\label{subsec:SOTA}

\setlength{\tabcolsep}{4pt}
\begin{table}[t]
\begin{center}
\caption{Comparison with state-of-the-art methods on PASCAL-Context.}
\label{table:sota_Pascal}
\scalebox{0.9}{
\begin{tabular}{l|ccccc|c}
 Method & Edge $\uparrow$ & SemSeg $\uparrow$ & Parts $\uparrow$ & Normals $\downarrow$ & Sal $\uparrow$ & $\Delta _m \%$ $\downarrow$ \\
\hline
Single-task & 71.88 & 66.22 & 59.69 & 13.64 & 66.62 & - \\
\hline
ASTMT (R-18 w/o SE) \cite{maninis2019attentive} & 71.20 & 64.31 & 57.79 & 15.06 & 66.59 & 3.49 \\
ASTMT (R-26 w SE) \cite{maninis2019attentive} & 71.00 & 64.61 & 57.25 & 15.00 & 64.70 & 4.12  \\
Series RA \cite{rebuffi2017learningRA1} & 70.62 & 65.99 & 55.32 & 14.27 & 66.08 & 2.97 \\
Parallel RA \cite{rebuffi2018efficientRA2} & 70.84 & 66.51 & 56.56 & 14.16 & 66.36 & 2.09 \\
RCM (ours) & 71.34 & 65.70 & 58.12 & 13.70 & 66.38 & \textbf{0.99} \\
\end{tabular}
}
\end{center}
\end{table}
\setlength{\tabcolsep}{1.4pt}

\setlength{\tabcolsep}{4pt}
\begin{table}[t]
\begin{center}
\caption{Comparison with state-of-the-art methods on NYUD. }
\label{table:sota_NYUD}
\scalebox{0.9}{
\begin{tabular}{l|cccc|c}
 Method & Edge $\uparrow$ & SemSeg $\uparrow$ & Normals $\downarrow$ & Depth $\downarrow$ & $\Delta _m \%$ $\downarrow$ \\
\hline
Single-task  & 68.83 & 35.45 & 22.20 & 0.56 & - \\
\hline
ASTMT (R-18 w/o SE) \cite{maninis2019attentive} & 68.60 & 30.69 & 23.94 & 0.60 & 6.96 \\
ASTMT (R-26 w SE) \cite{maninis2019attentive} & 73.50 & 30.07 & 24.32 & 0.63 & 7.56 \\
Series RA \cite{rebuffi2017learningRA1} & 67.56 & 31.87 & 23.35 & 0.60 & 5.88 \\
Parallel RA \cite{rebuffi2018efficientRA2} & 68.02 & 32.13 & 23.20 & 0.59 & 5.02 \\
RCM (ours) & 68.44 & 34.20 & 22.41 & 0.57 & \textbf{1.48} \\
\end{tabular}
}
\end{center}
\end{table}
\setlength{\tabcolsep}{1.4pt}

\begin{figure*}[t]
 \centering
   \subfloat[Input image]{\label{fig:in_img}
      \includegraphics[width=.32\textwidth]{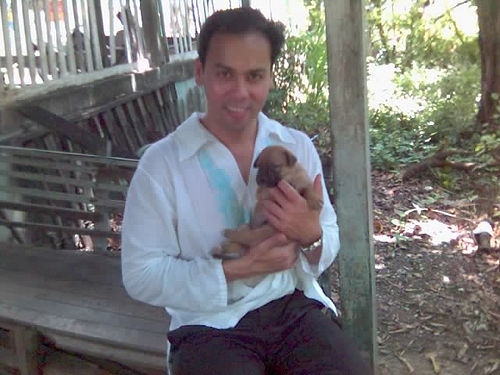}}
~
   \subfloat[Semseg]{\label{fig:semseg}
      \includegraphics[width=.32\textwidth]{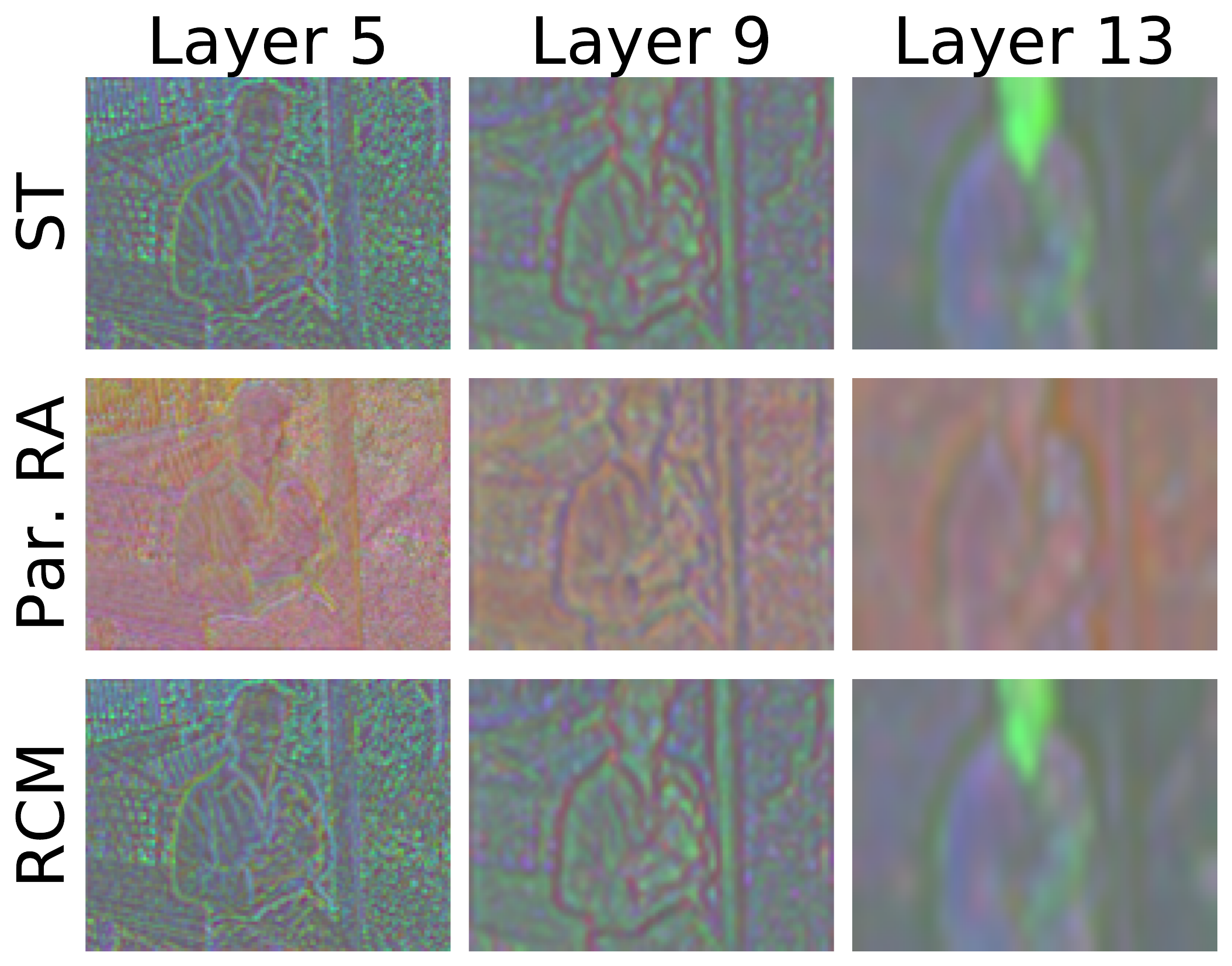}}
~
       \subfloat[Parts]{\label{fig:hum_parts}
      \includegraphics[width=.32\textwidth]{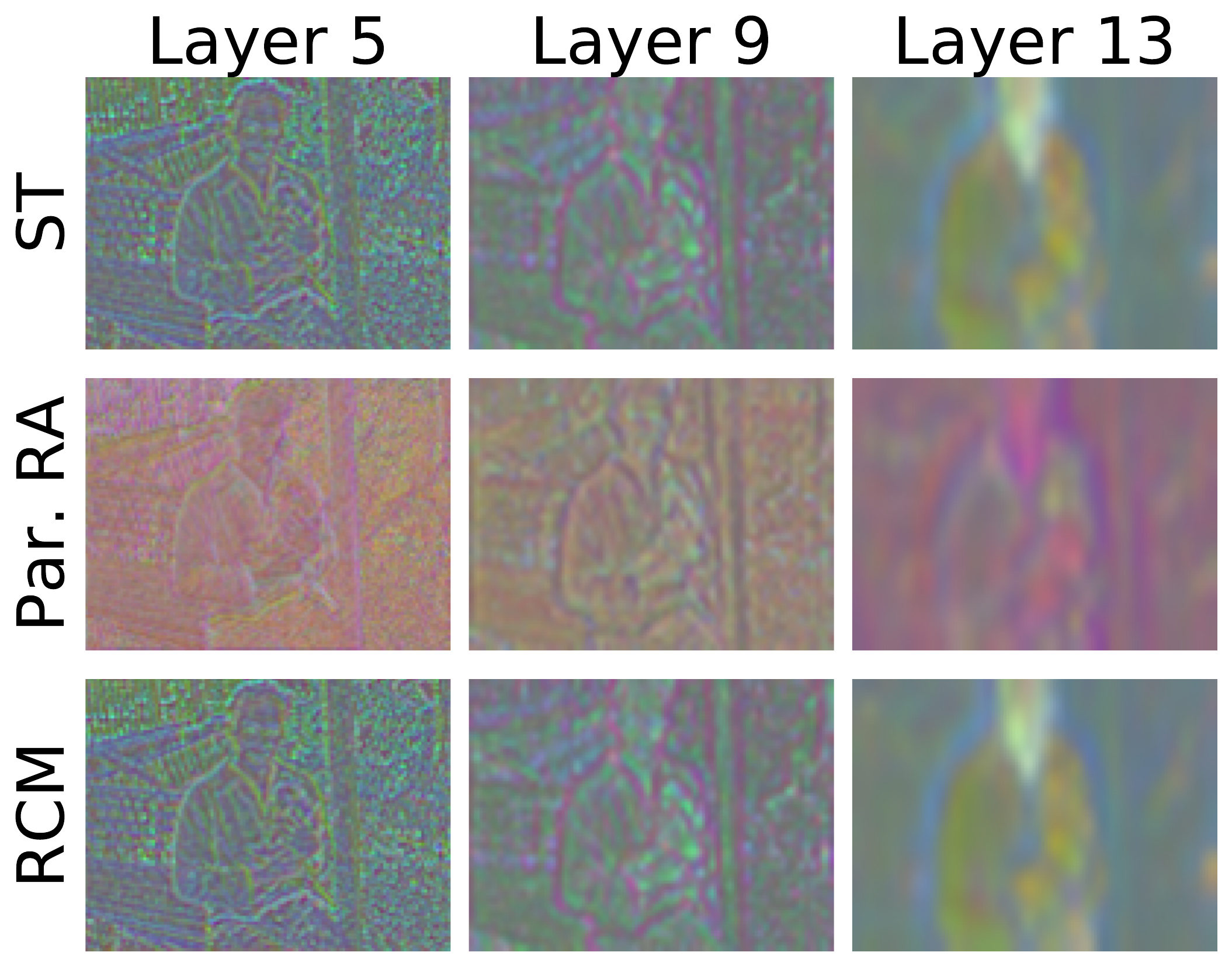}}

   \subfloat[Edge]{\label{fig:edge}
      \includegraphics[width=.32\textwidth]{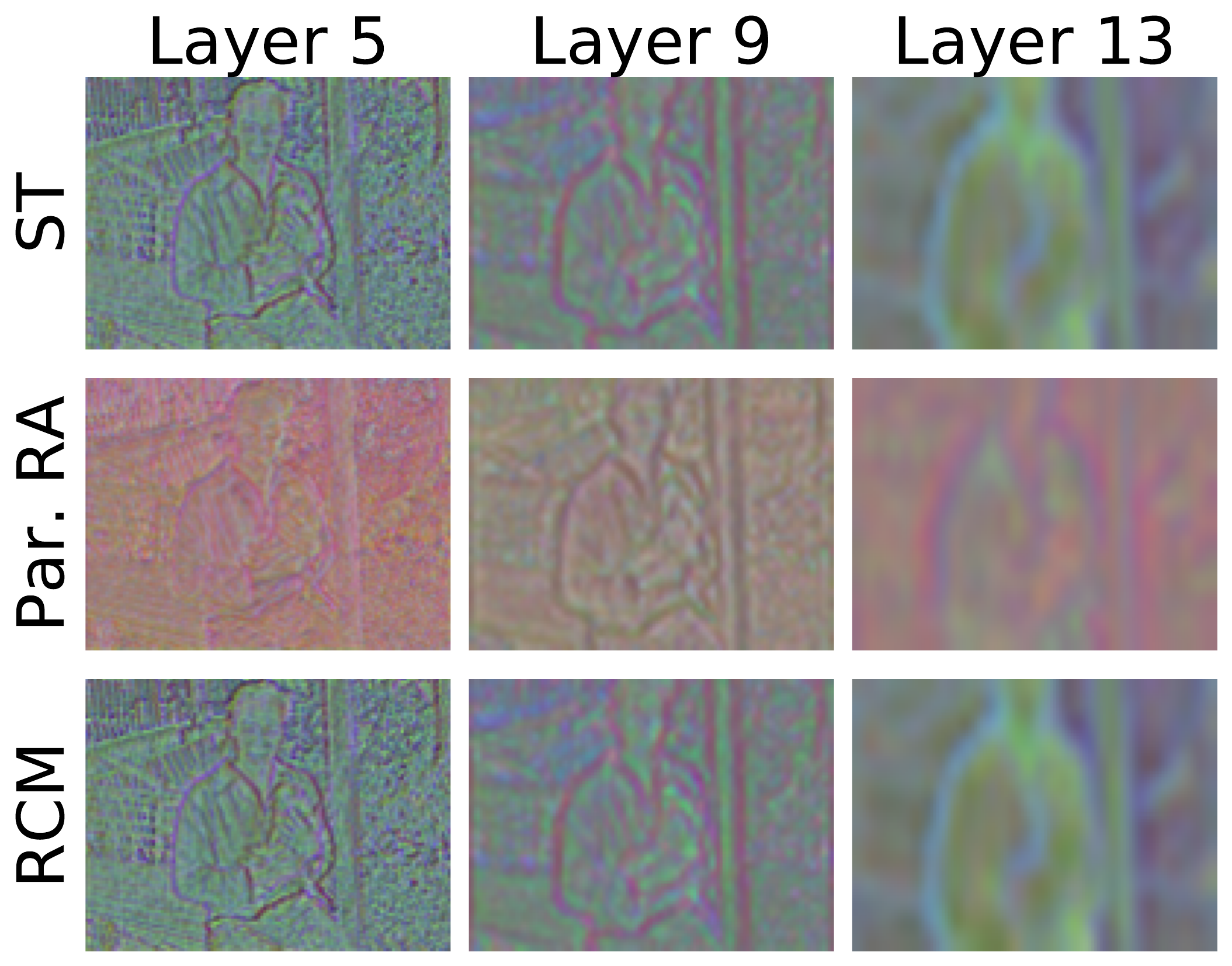}}
~
   \subfloat[Normals]{\label{fig:normals}
      \includegraphics[width=.32\textwidth]{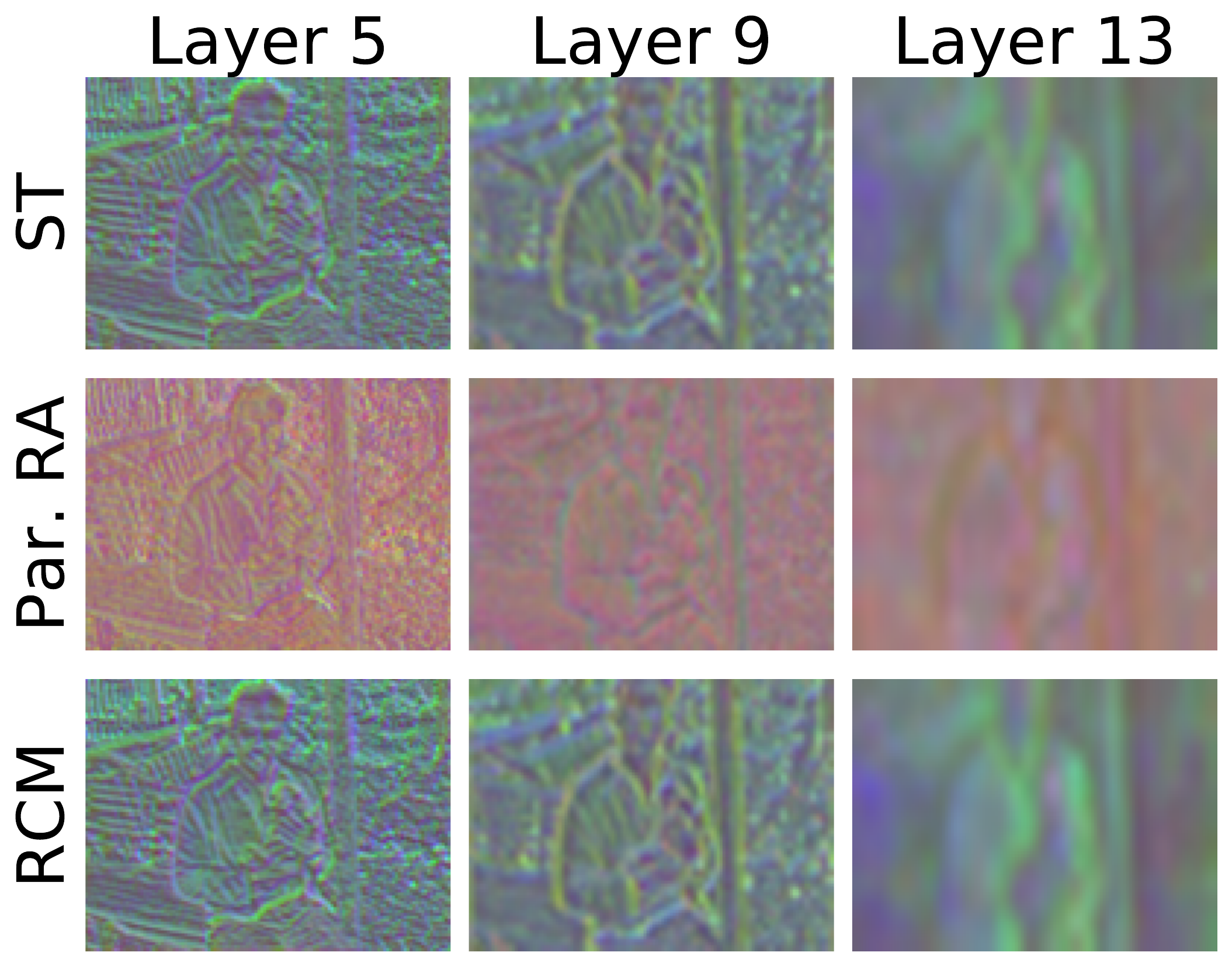}}
~
       \subfloat[Sal]{\label{fig:sal}
      \includegraphics[width=.32\textwidth]{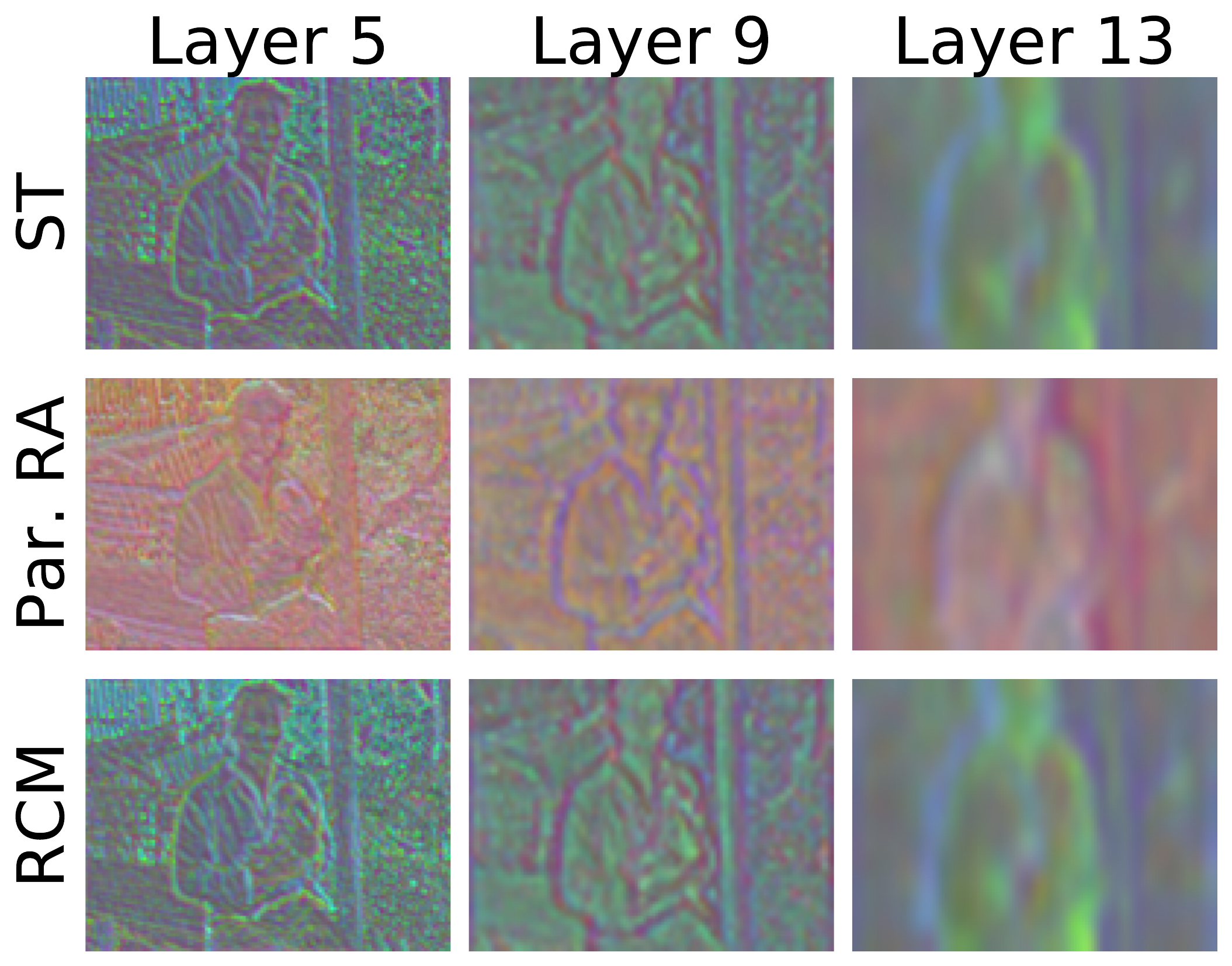}}
      
   \caption{\textbf{Feature visualizations.} We visualize the features of the input image (a) for the tasks of PASCAL-Context. The first row of each sub-figure corresponds to the responses of the single-task model (ST), the second row those of Parallel RA (Par. RA)~\cite{rebuffi2018efficientRA2} and the final row of our proposed method (RCM). For all tasks and depths of the network, the responses of RCM closely resemble those of ST, in contrast to the responses of Par.~RA. This is made apparent from the colours utilized by the different methods. The RGB values were identified from a common PCA basis across the three methods in order to highlight similarities and differences between them.}\label{fig:PCA}
   
\end{figure*}

In this work, we focus on comparing to task-conditional methods that can address MTL. We compare the performance of our method to Series Residual Adapter (Series RA) \cite{rebuffi2017learningRA1} and Parallel RA \cite{rebuffi2018efficientRA2}. Series and Parallel RAs learn multiple visual domains by optimizing domain-specific residual adaptation modules (rather than using RCM as in our work, Fig.~\ref{fig:RC_NFF}) on an ImageNet pre-trained backbone. Since both methods were developed for multi-domain settings, we optimize them using our own pipeline, ensuring a fair comparison amongst the methods while additionally benchmarking the capabilities of multi-domain methods in a multi-task setup. We further report the performance of ASTMT \cite{maninis2019attentive}, which utilizes an architecture resembling that of Parallel RA \cite{rebuffi2018efficientRA2} with Squeeze-and-Excitation (SE) blocks \cite{hu2018squeeze} and adversarial task disentanglement of gradients. Specifically, we report the performance of the models using a ResNet-26 (R-26) DeepLab-V3+ with SE as reported in \cite{maninis2019attentive}, and also optimize with the use of their codebase a ResNet-18 model without SE. The latter model uses an architecture resembling more closely that of the other methods since SE can be additionally incorporated in the others as well. We report the average performance drop with respect to our single-task baseline.

The results for PASCAL-Context (Table~\ref{table:sota_Pascal}) and NYUD (Table~\ref{table:sota_NYUD}) demonstrate that our method achieves the best performance, outperforming the other methods that make use of RA modules. This demonstrates that although the RA can perform competitively in multi-domain settings, placing the convolution in series without non-linearity is a more promising direction for the drastic adaptations required for different tasks in a multi-task learning setup.

We visualize in Fig.~\ref{fig:PCA} the learned representations of single-task, Parallel RA~\cite{rebuffi2018efficientRA2}, and RCM across tasks and network depths. For each task and layer combination, we compute a common PCA basis for the methods above and depict the first three principal components as RGB values. For all tasks and layers of the network, the representations of RCM closely resemble those of the single-task models. Simultaneously, Parallel RA is unable to adapt the convolution behavior to the extent required to be comparable to single-task models.

\subsection{Incremental learning for multi-tasking}
\label{subsec:IL}

\setlength{\tabcolsep}{4pt}
\begin{table}[t]
\begin{center}
\caption{Incremental learning experiments on a network originally trained on the low-level tasks (Edge and Normals) of PASCAL-Context.}
\label{table:IL_low_level_tasks}
\scalebox{0.9}{
\begin{tabular}{l|cc|ccc|c}
 Method & \textcolor{gray}{Edge $\uparrow$} & \textcolor{gray}{Normals $\downarrow$} & SemSeg $\uparrow$ & Parts $\uparrow$ & Sal $\uparrow$ & $\Delta _m \%$ $\downarrow$ \\
\hline
Single-task  & \textcolor{gray}{71.88} & \textcolor{gray}{13.64} & 66.22 & 59.69 & 66.62 & - \\
\hline
ASTMT (R-18 w/o SE) \cite{maninis2019attentive} & \textcolor{gray}{70.70} & \textcolor{gray}{14.84} & 55.32 & 50.49 & 64.34 & 11.77 \\
Series RA \cite{rebuffi2017learningRA1}  & \textcolor{gray}{70.62} & \textcolor{gray}{14.27} &  65.99 & 55.32 & 66.08 & 2.83 \\
Parallel RA \cite{rebuffi2018efficientRA2}  & \textcolor{gray}{70.84} & \textcolor{gray}{14.16} &  66.51 & 56.56 & 66.36 & 1.73 \\
RCM (ours)  & \textcolor{gray}{71.34} & \textcolor{gray}{13.70} & 65.70 & 58.12 & 66.38 & \textbf{1.26} \\
\end{tabular}
}
\end{center}
\end{table}
\setlength{\tabcolsep}{1.4pt}

\setlength{\tabcolsep}{4pt}
\begin{table}[t]
\begin{center}
\caption{Incremental learning experiments on a network originally trained on the high-level tasks (SemSeg and Parts) of PASCAL-Context.}
\label{table:IL_high_level_tasks}
\scalebox{0.9}{
\begin{tabular}{l|cc|ccc|c}
 Method & \textcolor{gray}{SemSeg $\uparrow$} & \textcolor{gray}{Parts $\uparrow$} & Edge $\uparrow$ & Normals $\downarrow$ & Sal $\uparrow$ & $\Delta _m \%$ $\downarrow$ \\
\hline
Single-task & \textcolor{gray}{66.22} & \textcolor{gray}{59.69} & 71.88 & 13.64 & 66.62 & - \\
\hline
ASTMT (R-18 w/o SE) \cite{maninis2019attentive} & \textcolor{gray}{63.91} & \textcolor{gray}{57.33} & 68.67 & 14.12 & 64.43 & 3.76 \\
Series RA \cite{rebuffi2017learningRA1}  & \textcolor{gray}{65.99} & \textcolor{gray}{55.32} &  70.62 & 14.27 & 66.08 & 2.39 \\
Parallel RA \cite{rebuffi2018efficientRA2}  & \textcolor{gray}{66.51} & \textcolor{gray}{56.56} & 70.84 & 14.16 & 66.36 & 1.88 \\
RCM (ours) & \textcolor{gray}{65.70} & \textcolor{gray}{58.12} & 71.34 & 13.70 & 66.38 & \textbf{0.52} \\
\end{tabular}
}
\end{center}
\end{table}
\setlength{\tabcolsep}{1.4pt}

We further evaluate the methods from Section~\ref{subsec:SOTA} in the incremental learning (IL) setup. In other words, we investigate the capabilities of the models to learn new tasks without the need to be completely retrained on the entire task dictionary. We divide the tasks of PASCAL-Context into three groups, \textbf{(i)} edge detection and surface normals (low-level tasks), \textbf{(ii)} saliency (mid-level task) and \textbf{(iii)} semantic segmentation and human parts segmentation (high-level tasks). IL experiments are conducted by allowing the base network to initially use knowledge from either (i) or (iii), and reporting the capability for the optimized model to learn additional tasks without affecting the performance of the already learned tasks (the performance drop is calculated over the new tasks that were not used in the initial training). In the IL setup, ASTMT~\cite{maninis2019attentive} is initially trained using an R-18 backbone without SE (a comparable backbone to the competing methods for a fair comparison) on the subset of the tasks (either i or iii). New tasks can be incorporated by training their task-specific modules independently. On the other hand, Series RA, Parallel RA, and RCM, were designed to be inherently incremental due to directly optimizing only the task-specific modules. Consequently, their task-specific performance in the IL setup is identical to that reported in Section~\ref{subsec:SOTA}.

In Tables~\ref{table:IL_low_level_tasks} and \ref{table:IL_high_level_tasks} we report the performance of tasks that are utilized to generate the initial knowledge of the model in grey (important for ASTMT~\cite{maninis2019attentive}), while in black the performance of the incrementally learned tasks. As shown in both tables, and in particular Table~\ref{table:IL_low_level_tasks}, ASTMT does not perform competitively in the IL experiments. This observation further demonstrates the importance of utilizing generic filter banks that can be adapted based on the task-specific needs, in particular for IL setups. We consider research in generic multi-task filter banks to be a promising direction.

%% file: text/conclusion.tex
\section{Conclusion}
\label{sec:conclusion}

We have presented a novel method of a convolutional operation reparameterization and its application to training multi-task learning architectures. These reparameterized architectures can be applied on a multitude of different tasks, and allow the CNN to be inherently incremental, while additionally eliminating task interference, all by construction. We evaluate our model on two datasets and multiple tasks, and show experimentally that it outperforms competing baselines that address similar challenges. We further demonstrate its efficacy when compared to the state-of-the-art task-conditional multi-task method, which is unable to tackle incremental learning. \\

\noindent
\textbf{Acknowledgments.} This work was sponsored by Advertima AG and co-financed by Innosuisse. We thank the anonymous reviewers for their valuable feedback.

%% file: text/sup_impl_details.tex
\section{Implementation Details}
\label{sec:impl_details}

We based our implementation details on the work of~\cite{maninis2019attentive}, listed below for completeness. \\

\noindent 
\textbf{Generic hyperparamaters.}
All models are optimized using SGD with a learning rate 0.005, momentum 0.9, weight decay 0.0001, and the ``poly'' learning rate schedule~\cite{chen2017deeplab}. We use a single GPU with a minibatch of 8 images. The input images during training are augmented with random horizontal flips and random scaling in the range [0.5, 2.0] in 0.25 increments. The validity of these hyperparameters has already been tested in~\cite{maninis2019attentive}, and hence they are used in all our experiments too, in order to ensure fair comparisons amongst different methods.\\

\noindent 
\textbf{Dataset specific hyperparameters.}
PASCAL-Context~\cite{mottaghi2014role} models are trained for 60 epochs. The spatial size of the input images is 512$\times$512. NYUD~\cite{silberman2012indoor} models are trained for 200 epochs. The spatial size of the input images is 425$\times$560. Images of insufficient size are padded with the mean color. \\

\noindent 
\textbf{Task weighting and loss functions.}
As is common in multi-task learning (MTL), losses require careful loss weighting \cite{maninis2019attentive,vandenhende2020mti,kendall2018multi,sener2018multi}, where each loss is task-dependent. For edge detection, we optimize the binary cross-entropy (BCE) loss, scaled by 50. Due to the class imbalance between the edge and non-edge pixels, edge pixels are penalized with a weight 0.95, while non-edge pixels with a scale of 0.05, accommodating~\cite{kokkinos2015pushing,maninis2017convolutional}. For evaluation, we set the maximum allowed mislocalization of the optimal dataset F-measure (odsF)~\cite{martin2004learning} to 0.0075 and 0.011 for PASCAL-Context and NYUD, respectively, using the package of~\cite{pont2015supervised}. Semantic segmentation and human parts segmentation are optimized with cross-entropy loss, weighted by the factors of 1 and 2, respectively. Predictions of surface normals (normalized to unit vectors) and depth modalities are penalized using the $\mathcal{L}_1$ loss, scaled by 10 and 1, respectively. Saliency is optimized using the BCE loss, weighted by a factor of 5. 

%% file: text/sup_reparam.tex
\section{Reparameterization Details}
\label{sec:reparam}

In Section~3.3 of the main text (Response initialization, RI), we introduced the methodology for the generation of a better filter bank $W_{s}$ when compared to that directly learned by pre-training $W_{s}$ on ImageNet, and demonstrated improved performance when utilizing RI in Section 4. In this section, we present additional detail. 

Recall that we defined $\boldsymbol{y} = f(\boldsymbol{x}; W^{m}) = W^{m}\boldsymbol{x}$ the responses of a convolutional layer for an input tensor $\boldsymbol{x}$, where $W^{m} \in \mathbb{R}^{c_{out} \times k^{2}c_{in}}$ are the pre-trained ImageNet weights. We specify $Y \in \mathbb{R}^{c_{out} \times n}$ as a matrix containing $n$ responses of $\boldsymbol{y}$ with the mean vector $\overline{\boldsymbol{y}}$ subtracted. To generate the new filter bank, we first compute the  eigen-decomposition of the covariance matrix $YY^{T} = USU^{T}$ (using Singular Value Decomposition, SVD), where $U \in \mathbb{R}^{c_{out} \times c_{out}}$ is an orthogonal matrix with the eigenvectors on the columns, and $S$ is a diagonal matrix of the corresponding eigenvalues. We can now utilize $UU^{T}$ which acts as a method to project to ($U^{T}$) and from ($U$) a latent space. Thus, we can rewrite $\boldsymbol{y} = UU^{T} (\boldsymbol{y} - \overline{\boldsymbol{y}}) + \overline{\boldsymbol{y}}$, with the centering operation being of importance due to the space $UU^{T}$ being generated from centred responses. This gives rise to
\begin{align}
\boldsymbol{y} & = W^{m}\boldsymbol{x} = UU^{T} (W^{m}\boldsymbol{x} - \overline{\boldsymbol{y}}) + \overline{\boldsymbol{y}} \nonumber\\
\boldsymbol{y} & = UU^{T}W^{m}\boldsymbol{x} + (\overline{\boldsymbol{y}} - UU^{T} \overline{\boldsymbol{y}})\nonumber\\
\boldsymbol{y} & = W^i_t W_s\boldsymbol{x} + b
\end{align}
where $W_{t}^{i}$, initialized by $U$, represents the task-specific parameters optimized independently for each task $i$, and is implemented as a $1 \times 1$ convolution. The non-trainable shared parameters are defined as $W_{s}=U^{T}W^{m}$ and implemented as a $k \times k$ convolution, with $k$ being the filter size of $W^m$. The bias $b$ can be added to the running mean of the batchnorm following the convolution \cite{ioffe2015batch}.

%% file: text/sup_baseline.tex
\section{Baseline}
\label{sec:baseline}

\setlength{\tabcolsep}{4pt}
\begin{table}[t]
\begin{center}
\caption{\textbf{Single-task baseline comparison.} We report the single-task performance of the baseline implementations of~\protect\cite{maninis2019attentive,vandenhende2020mti} for similar architectures on PASCAL-Context. The arrow indicates the direction for better performance.}
\label{table:sup_architecture}
\begin{tabular}{l|ccccc}
 Method & Edge $\uparrow$ & SemSeg $\uparrow$ & Parts $\uparrow$ & Normals $\downarrow$ & Sal $\uparrow$ \\
\hline
ASTMT \cite{maninis2019attentive} & 70.30 & 63.90 & 55.90 & 15.10 & 63.90  \\
MTI-Net \cite{vandenhende2020mti} & 68.20 & 64.49 & 57.43 & 14.77 & 66.38  \\
Ours & \bf{71.88} & \bf{66.22} & \bf{59.69} & \bf{13.64} & \bf{66.62} \\
\end{tabular}
\end{center}
\end{table}

To ensure our re-implementation provides a stable baseline, Table~\ref{table:sup_architecture} compares the single-task performance of our implementation using a ResNet-18 based DeepLabv3+, the results from~\cite{vandenhende2020mti} using a ResNet-18 based FPN~\cite{lin2017feature}, and the results from~\cite{maninis2019attentive} who utilized a ResNet-26 based DeepLabv3+. We demonstrate that our single-task baseline outperforms both works on every task, and even though the numbers are not directly comparable due to minor implementation differences, it provides a verification of a strong baseline. 

%% file: text/sup_Exp_VGG.tex
\section{Additional Backbone Experiments}
\label{sec:backbone_exp}

We additionally compare the proposed RCM (Reparameterized Convolutions for Multi-task learning) with respect to the single-task performance on the DeepLabv3+ with the deeper ResNet34 (R-34)~\cite{he2016deep} backbone. Results for PASCAL-Context~\cite{mottaghi2014role} and NYUD~\cite{silberman2012indoor} can be seen in Table~\ref{table:sup_sota_Pascal} and Table~\ref{table:sup_sota_NYUD}, respectively. As seen, the percentage drops of $1.18\%$ and $1.77\%$ for PASCAL-Context and NYUD respectively are comparable to that of the ResNet18 backbone reported in the main paper.

\setlength{\tabcolsep}{4pt}
\begin{table}[t]
\begin{center}
\caption{Comparison with the single-task baseline on PASCAL-Context for a DeepLabv3+ with an R-34 backbone.}
\label{table:sup_sota_Pascal}
\begin{tabular}{l|ccccc|c}
 Method & Edge $\uparrow$ & SemSeg $\uparrow$ & Parts $\uparrow$ & Normals $\downarrow$ & Sal $\uparrow$ & $\Delta _m \%$ $\downarrow$ \\
\hline
Single-task & 73.63 & 69.34 & 62.96 & 13.39 & 67.49 & - \\
\hline
RCM (ours) & 72.87 & 69.11 & 61.41 & 13.71 & 67.69 & \textbf{1.18} \\
\end{tabular}
\end{center}
\end{table}
\setlength{\tabcolsep}{1.4pt}

\setlength{\tabcolsep}{4pt}
\begin{table}[t]
\begin{center}
\caption{Comparison with the single-task baseline on NYUD for a DeepLabv3+ with an R-34 backbone.}

\label{table:sup_sota_NYUD}
\begin{tabular}{l|cccc|c}
 Method & Edge $\uparrow$ & SemSeg $\uparrow$ & Normals $\downarrow$ & Depth $\downarrow$ & $\Delta _m \%$ $\downarrow$ \\
\hline
Single-task  & 70.13 & 37.39 & 21.47 & 0.54 & - \\
\hline
RCM (ours) & 69.50 & 36.19 & 21.70 & 0.55 & \textbf{1.77} \\
\end{tabular}
\end{center}
\end{table}
\setlength{\tabcolsep}{1.4pt}